\def\year{2021}\relax
\documentclass[letterpaper]{article} 
\usepackage{aaai21}  
\usepackage{times}  
\usepackage{helvet} 
\usepackage{courier}  
\usepackage[hyphens]{url}  
\usepackage{graphicx} 
\usepackage{natbib}  
\usepackage{caption} 
\usepackage{mathtools}
\usepackage{amsmath}
\usepackage{cases}
\usepackage{booktabs}
\usepackage{multirow}
\usepackage{tabularx}
\usepackage{subfiles}

\newcommand\blfootnote[1]{%
  \begingroup
  \renewcommand\thefootnote{}\footnote{#1}%
  \addtocounter{footnote}{-1}%
  \endgroup
}
\usepackage{array}
\newcolumntype{L}[1]{>{\raggedright\let\newline\\\arraybackslash\hspace{0pt}}m{#1}}
\newcolumntype{C}[1]{>{\centering\let\newline\\\arraybackslash\hspace{0pt}}m{#1}}
\newcolumntype{R}[1]{>{\raggedleft\let\newline\\\arraybackslash\hspace{0pt}}m{#1}}
\newcolumntype{Y}{>{\centering\arraybackslash}X}

\title{VECA : A Toolkit for Building Virtual Environments to Train and Test Human-like Agents}

\author{
    Kwanyoung Park,  
    Hyunseok Oh,  
    Youngki Lee$^*$ \\
}

\affiliations {
    Department of Computer Science and Engineering \\
    Seoul National University, South Korea \\
    william202@snu.ac.kr, ohsai@snu.ac.kr, youngkilee@snu.ac.kr
}

\begin{document}

\noindent 
\maketitle
\blfootnote{$^*$Corresponding author.}

\begin{abstract}
\begin{quote}
Building \emph{human-like agent}, which aims to learn and think like human intelligence, has long been an important research topic in AI.
To train and test human-like agents, we need an environment that imposes the agent to rich multimodal perception and allows comprehensive interactions for the agent, while also easily extensible to develop custom tasks. 
However, existing approaches do not support comprehensive interaction with the environment or lack variety in modalities. Also, most of the approaches are difficult or even impossible to implement custom tasks.
In this paper, we propose a novel VR-based toolkit, VECA, which enables building fruitful virtual environments to train and test human-like agents. In particular, VECA provides a humanoid agent and an environment manager, enabling the agent to receive rich human-like perception and perform comprehensive interactions.  
To motivate VECA, we also provide 24 interactive tasks, which represent (but are not limited to) four essential aspects in early human development: joint-level locomotion and control, understanding contexts of objects, multimodal learning, and multi-agent learning.
To show the usefulness of VECA on training and testing human-like learning agents, we conduct experiments on VECA and show that users can build challenging tasks for engaging human-like algorithms, and the features supported by VECA are critical on training human-like agents.

\end{quote}
\end{abstract}

\section{Introduction}

\begin{table*}
\begin{tabularx}{\textwidth}{|l|Y|Y|Y|Y|Y|Y|Y|Y|Y|}
\hline
Environments & 3D & Large-scale & Exten sible& Physics & FPP Vision & Audio & Tactile & Multi-agent & Inter action
\\ \hline 
ALE~\cite{ALE}             & X  & X           & X            & X       & X          & X     & X       & X & X          
\\ \hline
DeepMind Lab~\cite{DeepMindLab}    & O  & X           & $\triangle$            & X       & O          & X     & X       & X & X
\\ \hline
OpenAI Universe~\cite{OpenAIUniverse} & O  & O           & O            & X       & O          & X     & X       & X & X 
\\ \hline
VizDoom~\cite{VizDoom}         & O  & X           & O            & X       & O          & X     & X       & O  & X       \\ \hline
Arena~\cite{Arena}           & O  & X           & O            & O       & X          & X     & X       & O  & X          \\ \hline
Malmo~\cite{Malmo}           & O  & O           & O            & X       & O          & X     & X       & X  & O        
\\ \hline 
Gibson~\cite{GibsonEnv}          & O  & O           & X            & X       & O          & X     & X       & X  & X           \\ \hline
MINOS~\cite{MINOS}           & O  & O           & X            & X       & O          & X     & X       & X  & X           \\ \hline
House3D~\cite{House3D}         & O  & O           & $\triangle$            & X       & O          & X     & X       & X   & X          \\ \hline
HoME~\cite{HoME}            & O  & O           & $\triangle$            & O       & O          & O     & X       & O   & X          \\ \hline 
AI2-THOR~\cite{AI2-THOR}        & O  & X           & O           & O       & O          & X     & X       & O  & O          \\ \hline \hline
VECA           & O  & X           & O            & O       & O$^+$          & O$^+$     & O       & O    & O            \\ \hline
\end{tabularx}
\caption{Comparison of various virtual environments with VECA. O$^+$ indicates that the environment supports the perception and also it reflects the characteristic of human perception.
\textbf{3D} : Supports 3D environment.
\textbf{Large-scale} : Number of environments is more than $10^3$.
\textbf{Extensible} : User can implement and add novel environments and tasks. $\triangle$ indicates that the user is possible, but difficult to add novel environment and tasks due to limitations such as lack of visual editor or heavily specialized API.
\textbf{Physics} : Supports physical properties (e.g. collision, friction)
\textbf{FPP Vision} : Renders first person perspective vision. 
\textbf{Audio} : Renders audio perception.
\textbf{Tactile} : Renders tactile perception.
\textbf{Multi-agent} : Supports multiple agents in single environment.
\textbf{Interaction} : Supports comprehensive interactions. For example in Malmo, hitting with a pickaxe breaks the block when it targets block while it gives damage when it targets monsters.
}
\end{table*}
It has long been an important research topic to understand how humans learn and build human-like agents~\cite{BuildingMachine,HumanLikeAgent}. In particular, human intelligence has been a role model for many modern learning machines as an interpretable and data-efficient general intelligence. For example, deep learning methods, inspired by the human brain structure, have outperformed previous state-of-the-art algorithms and even human intelligence in various domains such as image classification~\cite{ImagenetSurpass}, complex games such as Go or Starcraft 2~\cite{AlphaGo,Starcraft2}. Although there is no need to precisely duplicate human intelligence (which is error-prone and imperfect), human intelligence is still an attractive target to learn and get inspiration on how learning works~\cite{DeepLearningCritical}. 

Human intelligence learns by experience; collecting \textbf{rich multimodal perception} (such as vision, audio, tactile) from an environment~\cite{ObjectPerception,PerCog} and \textbf{actively interacting}~\cite{Active1,LearningPlay} with it. 
For instance, developmental psychologists have studied that general understanding of objects develops in an early stage of the toddler without any supervision, by receiving multimodal feedback during interaction on objects such as mouthing, chewing, and rotating~\cite{Gibson,Piaget}.
Like human intelligence, it has also been argued that artificial intelligence could benefit from multimodal~\cite{de2017guesswhat,ngiam2011multimodal} and interactive~\cite{caselles2019symmetry,hermann2017grounded} learning. 
To build agents that learn like human, it is crucial to provide rich multimodal perception and interactions for the agent.

Although prior works reveal that human-like agents learn from rich multimodal perception and active interaction with the environment, we still lack an understanding of \emph{how} and \emph{what} the agent learns. In particular, there are important unexplored problems, such as how human-like agents should be evaluated, and through what tasks human-like agents are trained. Such limitations motivate to develop an extensible toolkit, rather than a fixed set of tasks, to support the research community to train and test novel approaches and algorithms for human-like agents.

However, it is challenging to design an environment that can provide the agent with rich multimodal perception and comprehensive interactions while also being extensible to develop custom tasks. An intuitive way would be to develop a robot with an actionable body and sensors, but building such a human-like robot is highly costly. Also, it is difficult to test premature agents that may break robot hardware. Moreover, accelerating the training process is not straightforward since the robot has to perceive and interact in the real world. Another promising approach is to train the agent in a virtual environment. Existing environments such as game environments for reinforcement learning ~\cite{ALE,OpenAIUniverse,DeepMindLab,Malmo,VizDoom,Arena} have diverse games to evaluate the agent with some of the platforms are extensible. However, those environments oversimplify dynamics and perceptions of game avatars, which makes them inappropriate for developing human-like agents. There have been recent efforts to simulate realistic indoor environments ~\cite{GibsonEnv,AI2-THOR,HoME,House3D,MINOS}, but they do not support active interactions or lack of rich multimodal perceptions, and also hard or even impossible to implement custom tasks. 

In this light, we propose a novel VR-driven toolkit, VECA, which enables to train and test emerging human-like agents in a virtual environment. VECA provides essential features to train human-like agents: 1) rich human-like perceptions, 2) comprehensive interaction capability with the environment, 3) extensibility for implementing various custom tasks. 
Using VECA, developers can easily create a custom environment where an agent can take rich sensory inputs, learn cognition while interacting with the environment, and perform necessary actions to solve complex tasks.

More specifically, VECA is implemented over the \emph{Unity} engine, which provides not only realistic physics and rendering but also an user-friendly visual editor to the users. VECA provides a humanoid agent equipped with a humanoid avatar, which receives human-like perception and supports joint-level physical actions, as well as animation-based actions. To accurately simulate the agent's perception and action, VECA internally simulates the environment using low-level APIs of Unity instead of using the default simulation loop of Unity. Moreover, VECA provides a network-based python API, which enables training python-coded agents in VECA from external servers.

To motivate and formulate VECA, we showcase a set of environments equipped with 24 tasks for human-like agents, which reflect essential features in human learning (e.g., understanding contexts of objects, multimodal learning, multi-agent learning, joint-level locomotion and control). Those environments can be directly used or modified to train and test human-like learning algorithms.

We show the usefulness of VECA on training and testing human-like learning agents with various use cases. In particular, we analyze the results on the proposed tasks with widely used reinforcement learning algorithms and study the effect of the quality of perceptions. Our analysis shows that the performance across various tasks has noticeable differences, varying from easily solvable tasks to challenging tasks. We also show that the agent's performance can be reduced by $50\%$, $97\%$ if the spatialization, stereo feature of audio perception is removed, and by $20\%$ when the tactile perception is removed. Those results show that users can build challenging tasks for engaging human-like algorithms using VECA, and features of perceptions supported by VECA are critical to building environments for human-like agents.

The contribution of this work can be summarized as follows:

\begin{itemize}
\item {We propose a novel VR-based toolkit named VECA to train and test human-like agents in a virtual environment. VECA is the very first tool that provides rich human-like perceptions and interaction capability with a human avatar, which will serve as the cornerstone to develop innovative human-like models and algorithms. }

\item {We provide a network-based python API, which can be used to train agents from external servers without graphic devices.}

\item {We provide a set of tasks, datasets, and playgrounds, which can be directly used or modified to train and test various human-like learning algorithms.}

\item {By conducting various experiments on VECA, we show that users can build challenging tasks for engaging human-like algorithms using VECA, and features of VECA are important for training human-like agents. }
\end{itemize}


\section{Background \& Related Works}

Researchers have used various ways to train artificial intelligence: (1) datasets, (2) real robots, (3) virtual environments. However, they are not suitable for developing human-like agents, motivating us to build VECA to promote research on next-generation human-like agents. 

\subsubsection{Pre-collected datasets.}
The most common approach to train an agent is to use pre-collected datasets such as \emph{Imagenet}~\cite{Imagenet}, \emph{Audioset}~\cite{AudioSet}. 
However, this approach requires developers to collect a large volume of data to build accurate models, which is costly and time-consuming. Moreover, this approach is inappropriate for training agents that learn by interaction. 

\subsubsection{Training in reality.}
Another approach is to use a robot and train it in the real world~\cite{iCub, iCubTactile}. However, building robot hardware requires significant time, effort, monetary cost.
Furthermore, training agents in reality makes the training process hard to parallelize or accelerate, since resources and time scale of training process are bounded to those of the real-world. Those difficulties motivate the usage of virtual environments.

\subsubsection{Game-based environments.}
Recently, several game-based environments~\cite{ALE,OpenAIUniverse,DeepMindLab,Malmo,VizDoom,Arena} have been proposed and adopted to train and test agents with various learning algorithms (e.g., reinforcement learning~\cite{PPO}, imitation learning~\cite{GAIL}). 
However, tasks, dynamics, and perceptions from such game environments are oversimplified and do not apply to real-world problems.

\subsubsection{Realistic Indoor Simulators.}
To overcome the problems of game-based environments, realistic virtual environments such as
\emph{AI2-THOR}~\cite{AI2-THOR}, \emph{MINOS}~\cite{MINOS}, \emph{House3D}~\cite{House3D}, \emph{Gibson}~\cite{GibsonEnv}, \emph{HoME}~\cite{HoME}
have been proposed. 
On top of those, VECA aims to take one step further by including critical points in human learning.

Among those, the most related environments with VECA is \emph{HoME} and \emph{AI2-THOR}. \emph{HoME} provides a large-scale multimodal environment, which renders audio perception and enables physical interactions. However, it lacks tactile perception and diversity of interactions (only physical interaction). Also, it is hard to implement novel tasks due to lack of visual editor. \emph{AI2-THOR} supports object-specific interactions, which enables the agent to experience diverse circumstances. However, it does not support various modalities. Moreover, those environments do not reflect human characteristics, since those are not designed for this purpose. In short, VECA aims to support human-like modalities and diverse interactions, also extensible for easily implementing custom tasks.

\section{VECA Toolkit}

\begin{figure}
    \centering
    \includegraphics[width=0.45\textwidth]{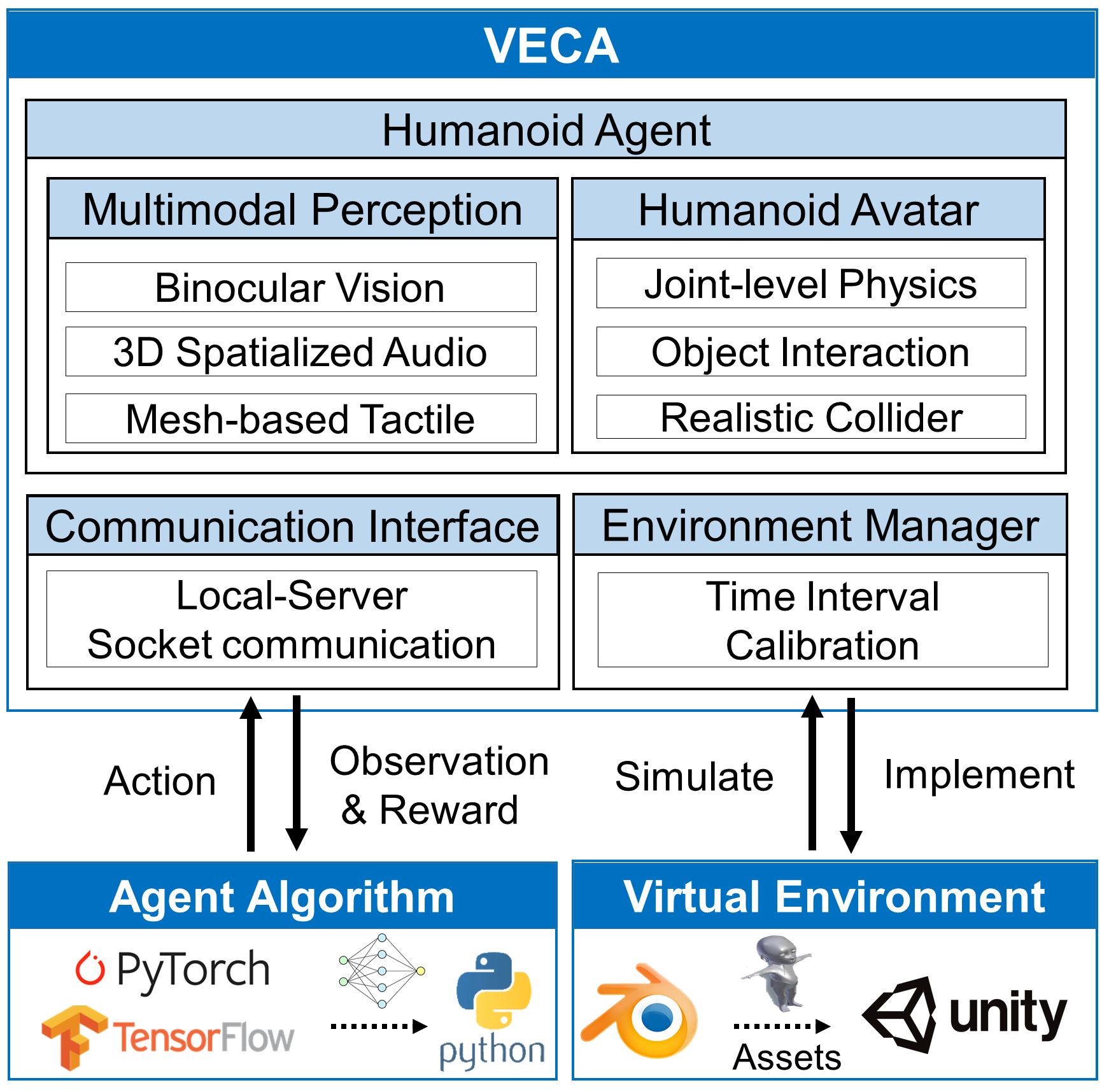}
    \caption{VECA architecture.}
    \label{fig:overview}
\end{figure}

VECA toolkit provides useful features for developing virtual environments to train and test human-like agents.
As shown in Fig. \ref{fig:overview}, VECA provides a \emph{humanoid agent} and an \emph{environment manager}, enabling the agent to receive rich human-like perception and perform comprehensive interactions.
To provide an extensible toolkit for the users, we chose \emph{Unity} as an engine to simulate and develop the environment, a state-of-the-art game engine equipped with realistic rendering, physics, and user-friendly visual editor. On top of that, users can train and test python-coded agents from external servers using network-based \emph{communication interface} provided by VECA.


\subsection{Humanoid Agent}
This component simulates an actionable agent with multimodal perception, which can be trained by human-like learning algorithms. The agent interacts with the virtual environment through its \emph{humanoid avatar}, capable of performing joint-level actions and pre-defined actions such as walking, kicking. The agent receives four main perceptions: vision, audio, tactile, and proprioception (including vestibular senses), which can be perceived with various configurations.  

\subsubsection{Humanoid Avatar \& Action.}
We modeled a humanoid avatar with an accurate collider and human-like joint-level motion capability, which allows the agent to perform joint-level actions and physically interact with the environment. 
Although there are many humanoid assets in Unity, they are mainly designed for games and do not precisely model bodies and actions.
In this light, we model a humanoid agent with Blender3D software and integrate it into a Unity asset, which is easy to customize.  
Specifically, the avatar has 47 bones with 82 degrees of freedom with hard constraints similar to human joints, and skin mesh represented with 1648 triangles, which adaptively changes according to the orientation of the bones.

However, training an agent from joint-level actions and physical interactions may not be practically plausible for complex tasks. To mitigate the user's burden of training those tasks, we also support an animation-based agent, which supports stable primitive actions (e.g., walk, rotate, crawl) and interactions (e.g., grab, open, step on) by trading off its physical plausibility. Specifically, we classify objects with their mass: For the light objects, the agent is relatively kinematic and apply the collision force only to the object. For a heavy object, the object is relatively kinematic and pushes the agent out of the collision. 

On top of that, VECA provides an user-friendly interface where the users can implement interactions between agent and object. 
Similar to \textit{AI2-THOR}~\cite{AI2-THOR}, the object is interactive with the agent when the object is visible (rendered by agent's vision and also closer than 1.5m) and not occluded by transparent objects. When there are multiple interactable objects, the agent could choose among those objects or follow the default order (closest to the center of the viewport). 

\subsubsection{Vision.} To mimic human vision, we implement a binocular vision using two Unity RGB cameras. The core challenge for vision implementation lies in simulating the diverse variants of human vision in the real-world. For instance, human vision varies with age: the infant's vision has imperfect color and sharpness~\cite{babyColor, babyBlur}. To simulate these various factors, we design multiple visual filters (e.g., changing focal length, grayscale, blur) and allow the users to use these features in combination to simulate a particular vision system.

\subsubsection{Audio.} The auditory sense allows human to recognize events in the blind spot and even roughly estimate the audio source's position. 
This is possible because the auditory sensor uses the time and impulse difference between two ears and receives the sound affected by its head structure ~\cite{HRTF}. A common approach to simulate hearing is to use \emph{AudioListener} provided by Unity and utilize the Unity SDK for 3D spatialization. However, we found that this approach makes it impossible to design tasks with multiple agents or to simulate multiple environments in a single application since Unity supports only one listener per scene. 

To provide human-like rich auditory sense to the agent, we address the problems in two folds. 
First, we adopt the LISTEN HRTF (head-related transfer function) dataset~\cite{LISTEN-HRTF} to simulate the spatial audio according to the human's head structure.
Second, we allow multiple listeners in the scene and supports various effects such as 3D spatialization or reverb. To support those features, each listener maintains the list of audio sources, calculates the distance from the agent and the room impulse~\cite{RIR} of the audio source, and simulates the audio using those pieces of information. Note that we also enable the developer to listen to the agent's audio data using the audio devices for debugging purposes.

\subsubsection{Tactile.} Tactile perception takes an important role when the agent physically controls and interacts with the object, but it is challenging to model in Unity.
A common approach to implement tactile perception is to model the body as a system of rigid bodies and calculate the force during the collision with other objects by dividing the collision impulse with the simulation interval.
However, this approach makes the tactile perception sensitive to the simulation interval since rigid body collision happens instantly. For example, if we halve the simulation interval for accurate simulation, then the collision force is doubled, although the same collision happened. 
Moreover, Unity only imposes the total impulse of the collision,
so the tactile data would be incorrectly calculated when there are multiple contact points, which frequently happens in real-life situations (e.g., hand grabbing a phone).

To simulate tactile perception, we approximate the collision force using Hooke's law, which has not only been adopted for implementing virtual tactile sensors~\cite{Tactile1} but also used for real tactile sensors~\cite{Tactile2}.
Specifically, we model the human body as rigid bones covered with flexible skin (which obeys Hook's law). When the collision occurs in the soft skin, we calculate the force using Hooke's law. If the collision occurs in the bone, the collision is handled by the physics engine and the force is calculated as the maximum force. We normalize the tactile input by dividing the force with the maximum force, which is calculated by Equation \ref{equation:tactile} with the spring displacement $d$ with the maximum displacement $d_{\max}$.

\begin{equation}
    T(d) = \min(1,\frac{d}{d_{\max}})
    \label{equation:tactile}
\end{equation}

For the position of sensors, we distribute six sensors on triangle formulation for each triangle in the mesh of the agent. Each sensor senses the normal component of the normalized pressure to its triangle using Eq. \ref{equation:tactile}.

\subsubsection{Proprioception.} VECA also provides proprioceptive sensory data. The raw forms of human proprioceptive sense, such as pressure sensors from muscles and joints, are hard to simulate and give unnecessary noise to the desired proprioceptive information. Thus, we directly provide physical information, such as bone orientation, current angle, and angular velocity of the joints, as proprioception.

\begin{figure*}
    \centering
    \includegraphics[width=\textwidth]{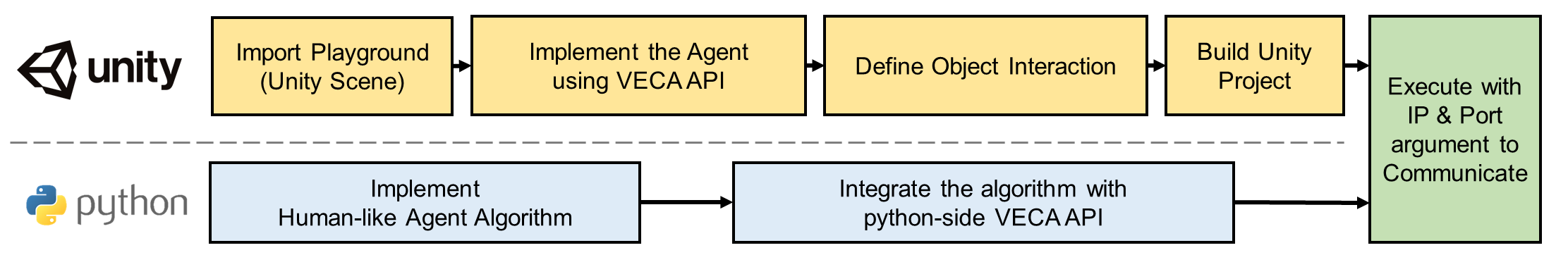}
    \caption{Workflow diagram about how VECA can be used in training and testing human-like agents. 
    On the environment side, the user designs tasks and implement environment using assets of Unity. Using VECA, the users can import humanoid agent with human-like perception and interaction, define interactions between object and agent, with no concern about management of the inner simulation loop for the environment.
    On the algorithm side, the user designs and implements novel human-like learning algorithms using python. Using VECA, the user can realize those algorithms with the environment.
    }
    \label{fig:Workflow}
\end{figure*}

\subsection{Environment Manager}

We design an environment manager that collects observations and performs actions with the avatar in the Unity environment. The main challenge is that the time interval between consecutive frames fluctuates in Unity. This is because Unity focuses on providing perceptually natural scenes to the user and adapt the frame rate to available computing power. Also, communication latency affects the time intervals; for instance, when there is a large communication delay, the simulation time is fast-forwarded to compensate for the delay. Although the simplest way would be to limit the users to implement the environment in \emph{FixedUpdate()} (which is called in a fixed rate), this approach makes \emph{Update()} function, which is mainly used in Unity, out of sync. 

To address the problem, we separate the clock of the virtual environment from the actual time by implementing the time manager class \emph{VECATime}, which replaces the \emph{Time} class in Unity. Specifically, we enable VECA to simulate time-dependent features (e.g., physics, audio clip, animation) with constant time steps. For the physics, VECA adopts the physics simulator provided by Unity to simulate physics for the constant time step. For other time-dependent features, VECA searches for all objects with corresponding features in the environment and explicitly controls the simulation time to enforce a constant time interval.

\subsection{Communication Interface}
This component manages the data flow between the agent algorithm (Python) and the environment (Unity), and provides a socket-based network connection to support training agents on remote servers. Since Unity does not support rendering visual information in servers without graphics devices, the Unity environment needs to be executed in a local machine with displays to get visual data or visualize the training status. This component allows the communication between the agent in the remote server with the Unity environment in the local machine. 

\begin{figure*}
    \centering
    \includegraphics[width=\textwidth]{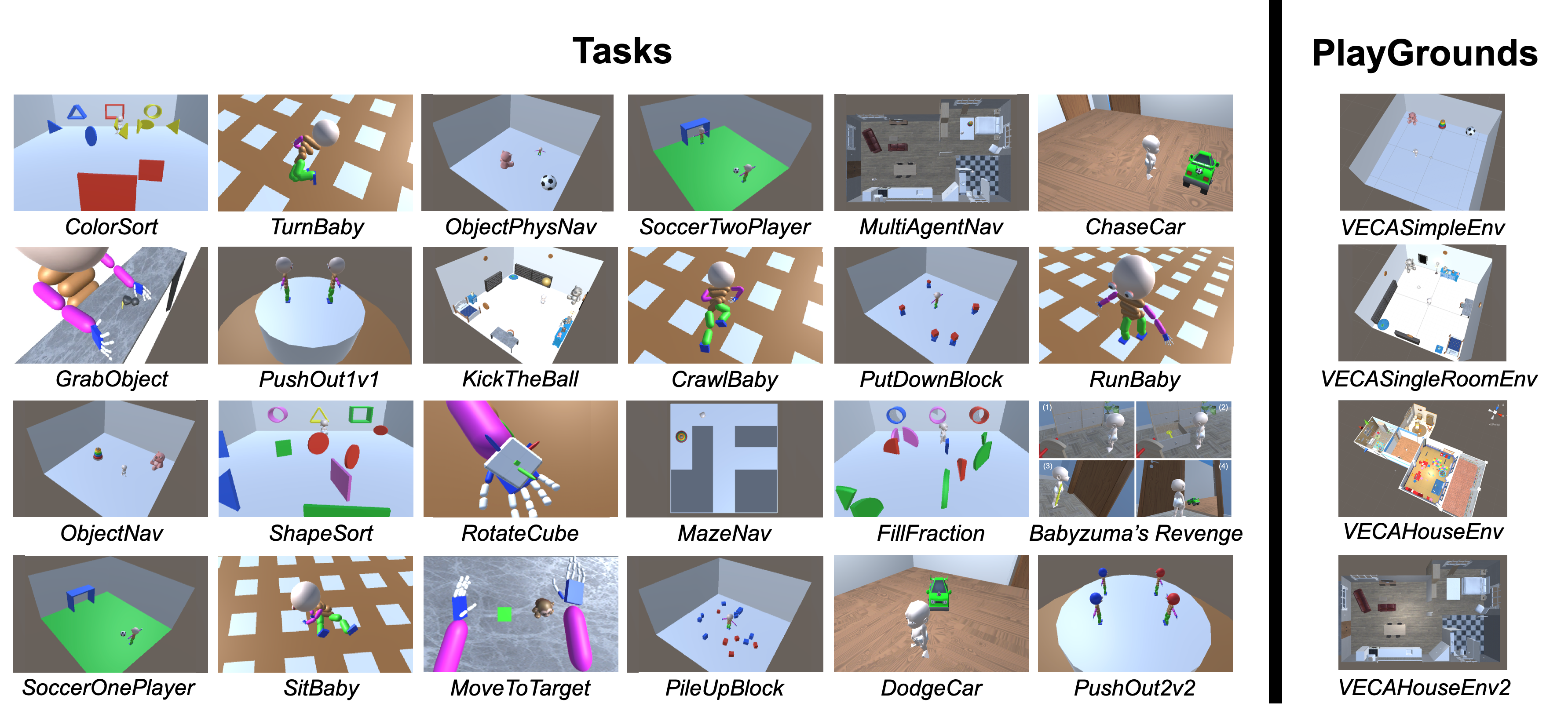}
    \caption{Set of tasks and playgrounds supported by VECA.}
    \label{fig:environments}
\end{figure*}

\subsection{How to Use VECA}

VECA provides user-friendly interfaces to design custom environments, tasks, agents, and object-agent interactions as shown in Fig. \ref{fig:Workflow}. Firstly, users need to create a virtual environment. For this, the user could rely on Unity, which has many existing assets and APIs to implement virtual environments. Secondly, users need to define the perceptions and actions of an agent. For this, VECA provides a humanoid avatar and useful APIs to customize it. In specific, users need to implement three main functions (similar to ML-Agents~\cite{ML-Agents}): (1) AgentAction(action) to make the agent perform a particular action, (2) CollectObservation() to collect various sensory inputs, and (3) AgentReset() to reset the agent. To be more specific, users can implement AgentAction(action) by importing existing interactions or implementing new interactions over \emph{VECAObjectInteract} interface. Finally, the user builds the environment (including the agent) as a standalone application. Those environments could be used to train the agents with novel learning algorithms in python, where many machine learning libraries exist, using the python API provided by VECA.

\section{Various Tasks Towards Human Intelligence}

To motivate VECA, we provide various tasks that represent building blocks of human learning, as shown in Fig \ref{fig:environments}. We focused on four essential aspects in early human development: joint-level locomotion and control, understanding contexts of objects, multimodal learning, and multi-agent learning. Note that these categories are neither disjoint nor unique. 

\subsubsection{Joint-level Locomotion \& Control.} 
We provide a set of tasks that feature learning \emph{joint-level locomotion and object-humanoid interaction} with diverse objectives.
Humanoid agent control has been studied for human assistance and also to understand the underlying human cognition. However, developing a human-level control is extremely challenging~\cite{akkaya2019rubiksOpenAI}.
Prior environments are domain-specific and difficult to integrate the human-like aspects.
VECA supports joint-level dynamics down to finger knuckles, physical interactions, and tactile perception. With this capability, we build a set of challenging joint-level control tasks \{TurnBaby, RunBaby, CrawlBaby, SitBaby, RotateCube,  SoccerOnePlayer\}. We also incorporate multimodal learning tasks (GrabObject) and tasks related to understanding objects (PileUpBlock, PileDownBlock).

\subsubsection{Understanding Contexts of Objects.}
We provide various tasks \{ColorSort, ShapeSort, ObjectNav, BabyZuma's Revenge, MazeNav, FillFraction\} which aims to understand \emph{contexts of objects}. Learning abstract contexts of objects is one of the key features in human learning~\cite{ContextHuman1,ContextHuman2}, and is getting increasing attention in artificial intelligence~\cite{ContextAI1,ContextAI2}. Thanks to the extensibility of VECA, users can implement tasks and environments with various useful contexts, including properties (ColorSort, ShapeSort, ObjectNav), functionality (BabyZuma's Revenge), and mathematical meanings (FillFraction).

\subsubsection{ Multimodal Learning.} 
For multimodal learning, we provide tasks \{KickTheBall, ObjectPhysNav, MoveToTarget\} which features in learning how to incorporate \emph{multiple sensory inputs}. Multimodal learning plays a big role in human learning\cite{PerCog,ObjectPerception}, and have also been suggested to be beneficial for training artificial intelligence\cite{de2017guesswhat,ngiam2011multimodal}. Using rich perceptions provided by VECA, users can develop tasks and environments for multimodal learning, such as vision-audio (KickTheBall) and vision-tactile (ObjectPhysNav, MoveToTarget) learning.

\subsubsection{Multi-Agent Reinforcement Learning.} 
We provide a set of multi-agent RL tasks in which agents need to \emph{cooperate or compete with others} to solve the tasks. Multi-agent learning is critical in human development
~\cite{MultiAgentHuman1,MultiAgentHuman2}
and is a rising topic for artificial general intelligence
~\cite{MultiAgentAI1,MultiAgentAI2}. 
However, it is difficult to generate a multi-agent task and to integrate the 
multimodal perception and active interactions (including interactions between agents) 
to the task. With the multi-agent support of VECA, we build a competitive task (SoccerTwoPlayer), a cooperative task (MultiAgentNav), and a mixed competitive-cooperative task (PushOut). Some of them also include other features such as multimodality  (MultiAgentNav) or joint-level physics (SoccerTwoPlayer).

\subsubsection{Supervised Learning.} 

Although VECA is designed for interactive agents, supervised learning still plays an important role in training/testing machine-learning agents. For example, some users may pre-train their agent with supervised tasks or evaluate their agent using supervised downstream tasks. In this light, we also provide labeled datasets collected from VECA for several supervised learning problems, such as image classification, object localization, sound localization, and depth estimation. See the appendix for a detailed description of the datasets.

\subsection{Playgrounds}

VECA provides four exemplar playgrounds, as shown in Fig \ref{fig:environments}. Those playgrounds vary in numbers of props, furniture, and its structure. It allows controlling the complexity of tasks such as \emph{ObjectNav, KickTheBall, MultiAgentNav}, i.e., the same tasks could be tested in different complexity. For the details of the playgrounds, please refer to the appendix.

\section{Experiments}

\begin{figure*}
    \centering
    \includegraphics[width=\textwidth]{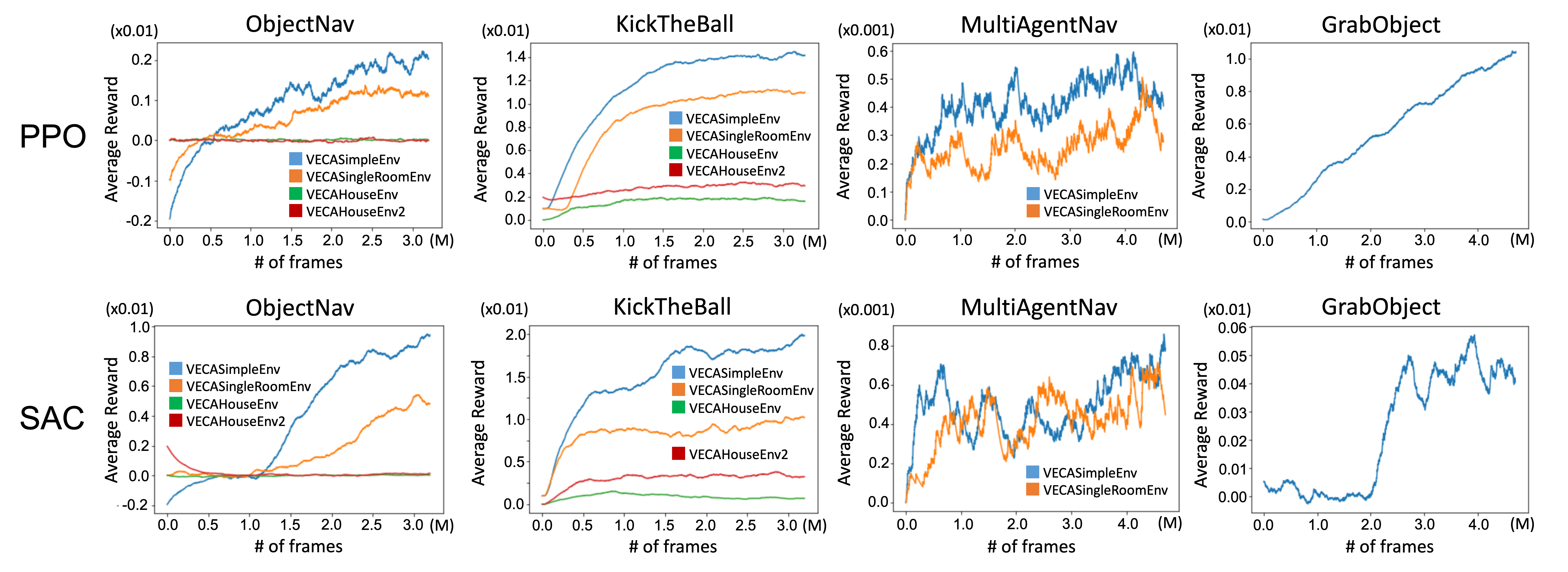}
    \caption{Visualization of training progress over average reward of agents trained by PPO (Proximal Policy Optimization) and SAC (Soft Actor-Critic). The agents are trained on subset of tasks provided by VECA: \emph{ObjectNav}, \emph{KickTheBall}, \emph{MultiAgentNav}, \emph{GrabObject}, which represents essential aspects in early human development. Best viewed in color.}
    \label{fig:result_all}
\end{figure*}

\begin{figure}
    \centering
    \includegraphics[width=0.48\textwidth]{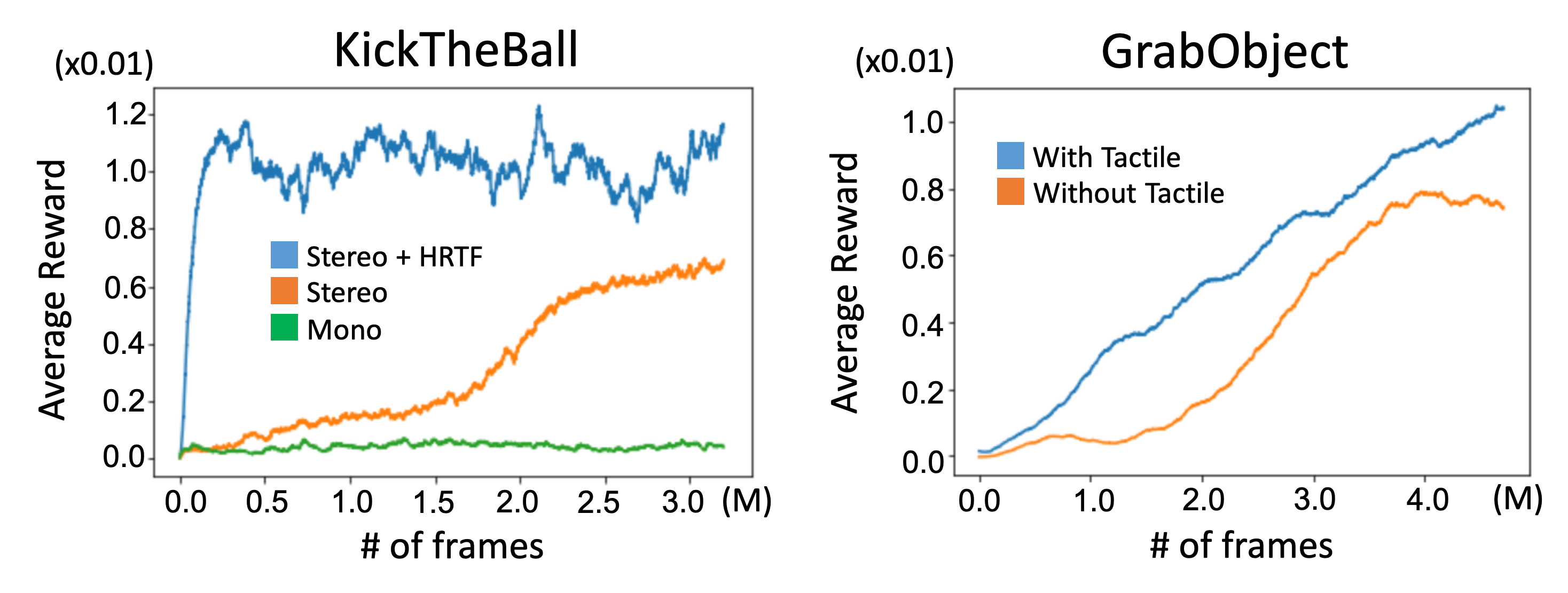}
    \caption{Learning curve (average reward by step) for \emph{KickTheBall} task with different quality of auditory perception and \emph{GrabObject} task with/without tactile perception. Best viewed in color.}
    \label{fig:resultPerception}
\end{figure}

We performed a set of experiments to show and evaluate the effectiveness of VECA on training and testing human-like agents. Please note that although it is ideal to evaluate the agents with human-like learning algorithms, we used conventional reinforcement/supervised learning algorithms since they are yet to be fully developed due to a lack of tools like VECA. 

\subsection{Results on Tasks for Human-like Agents}

\subsubsection{Experiment Setup.}
Among various RL algorithms, we applied PPO~\cite{PPO} and SAC~\cite{SAC}, widely used state-of-the-art reinforcement learning algorithms that can be used to learn VECA-supported example tasks. For the tasks, we picked one representative task per each aforementioned task categories: \emph{GrabObject}, \emph{ObjectNav}, \emph{KickTheBall}, \emph{MultiAgentNav}. We experimented with all compatible playgrounds to also observe the effect of playgrounds on the agent's performance. 

\subsubsection{Results.}
As shown in Fig. \ref{fig:result_all}, performance of learning algorithms show noticeable differences, varying from easily solvable tasks (\emph{ObjectNav}, \emph{KickTheBall}) to highly challenging tasks (\emph{MultiAgentNav}, \emph{ObjectNav} and \emph{KickTheBall} in \emph{VECAHouseEnv}, \emph{VECAHouseEnv2}). It shows that there is a room for novel human-like learning algorithms to strike in, and also indicates that using VECA, users can build challenging tasks for engaging human-like agents.

Also, for the \emph{KickTheBall} and \emph{ObjectNav} tasks, the performance of the agent and its training speed significantly differ with the playground. In detail, learning \emph{KickTheBall} task in the \emph{VECASimpleEnv} and \emph{VECASingleRoomEnv} playground shows noticeable performance enhancement, but the performance increase is negligible in the \emph{VECAHouseEnv} and \emph{VECAHouseEnv2} playground. It shows that the users can moderate the difficulty of the task using the playgrounds. 

\subsection{Effectiveness of Multimodal Perception}

\subsubsection{Experiment Setup.} We compared the performance of the agents trained with PPO in various quality of auditory and tactile perception, to show that the perception in VECA takes an essential role in training human-like agents. For auditory perception, we trained an agent to perform the \emph{KickTheBall} task in \emph{VECASingleRoomEnv} with varying audio quality: stereo + HRTF, stereo (without HRTF), and mono (without spatialization). For tactile perception, we trained an agent to perform the \emph{GrabObject} task trained with and without tactile perception. We evaluated the performance of the agent with the average reward of the agent. For technical details, please refer to the appendix.

\subsubsection{Results.} 
As shown in Fig. \ref{fig:resultPerception}, the existence and the quality of the perception play an important role in the agent's performance. On the auditory perception, the agent without spatialized audio got a low score since the agent couldn't use the audio information to figure out the direction of the ball. The agent without HRTF was able to use the information of audio to get a high score, but it required much more training time to learn it. The agent with full spatialized audio could get the maximum score with short training time. On the tactile perception, there exists a noticeable difference between agents with and without tactile perception in terms of final performance and its training speed. These results show that the features of perceptions provided by VECA are critical to the agent's performance.

\section{Conclusion}

We proposed VECA, a virtual toolkit for building virtual environments to train and test emerging \emph{human-like agents}. VECA provides a virtual humanoid agent with rich human-like perceptions, a joint-level physics, and an environment for the agent to interact, facilitating the development of the human-like agents.
VECA also provides an environment manager managing the internal simulation loop of the environment for the agent, and a communication interface, which enables the users to train the agent using python-based learning algorithms. 
Our experiments show that various tasks towards human intelligence can be easily generated with VECA, and they are challenging to solve with recent RL algorithms. Moreover, multimodal perception of VECA plays an important role in training human-like agents. We believe that the features of VECA would be useful for developing environments to train and test human-like agents.

\section{Acknowledgement}

This work was supported by Institute of Information \& Communications Technology Planning \& Evaluation (IITP) grant funded by the Korea government (MSIT) (No. 2019-0-01371, Development of brain-inspired AI with human-like intelligence).

\bibliography{VECAPaperAAAI21.bib}

\clearpage

\def\hlinewd#1{%
\noalign{\ifnum0=`}\fi\hrule \@height #1 %
\futurelet\reserved@a\@xhline} 

\section{Dataset for Supervised Learning}

We collected the data in \emph{VECASimpleEnv} to prevent multiple objects is included in the vision datasets. All dataset consists of 80000 image/audio/tactile data, collected in various perspectives, orientations.

\subsubsection{Object classification (Vision).}
The agent has to classify images by its included object. There are three types of objects: doll, ball, and pyramid. 

\subsubsection{Object classification (Tactile).}
The agent s to classify four shapes: pyramid, sphere, cube, cylinder. The object is dropped from above the hand. The agent receives the tactile data sensed by hand during 128 time-steps (total 0.512s). 

\subsubsection{Distance estimation (Vision).} The agent has to estimate the distance between the camera and the object. Note that the distance uses the meter unit, so it may not be practical to apply regression without normalization.

\subsubsection{Object recognition (Vision).} The agent has to estimate the bounding box of the object projected to the camera. The bounding box is represented as (x, y, h, w), which indicates x and y coordinate of the center, height, and width.

\subsubsection{Sound localization (Audio).}
The agent has to localize the direction of the sound. The agent receives spatialized audio data for 0.2s.

\subsection{Training Details \& Baseline results}
We used NVIDIA GeForce GTX 1060 6GB to train the agent in Ubuntu 16.04. Results are averaged with five runs. We used the architecture the same as those of reinforcement learning experiments. We trained the model for 50 epochs. Baseline results are shown in Table \ref{table:resultData}.

\begin{table}[h!]
\begin{tabular}{|c|c|}
\hline
Task(Perception)         & Result (metric)                 \\ \hline
Classification (Vision)  & 99.7\% $\pm$ 0.02\% (top-1 accuracy)         \\ \hline
Classification (Tactile) & 94.8\% $\pm$ 0.64\% (top-1 accuracy)         \\ \hline
Recognition (Vision)     & 76.1\% $\pm$ 1.52\% (top-1 accuracy)         \\ \hline
Distance (Vision)        & 1.90  $\pm$ 0.30 (relative abs. error) \\ \hline
Localization (Audio)     & 0.80\% $\pm$ 0.03 (cosine similarity)        \\ \hline
\end{tabular}
\caption{Baseline result of datasets.}
\label{table:resultData}
\end{table}

\section{Training Details: RL}
For our experiments, we used Intel(R) Core(TM) i5-9600KF with 3.70GHz for simulating the environment in Windows 10 and used NVIDIA GeForce GTX 1060 6GB to train the agent in Ubuntu 16.04. Results are obtained with single run.

\begin{figure}
    \centering
    \includegraphics[width=0.38\textwidth]{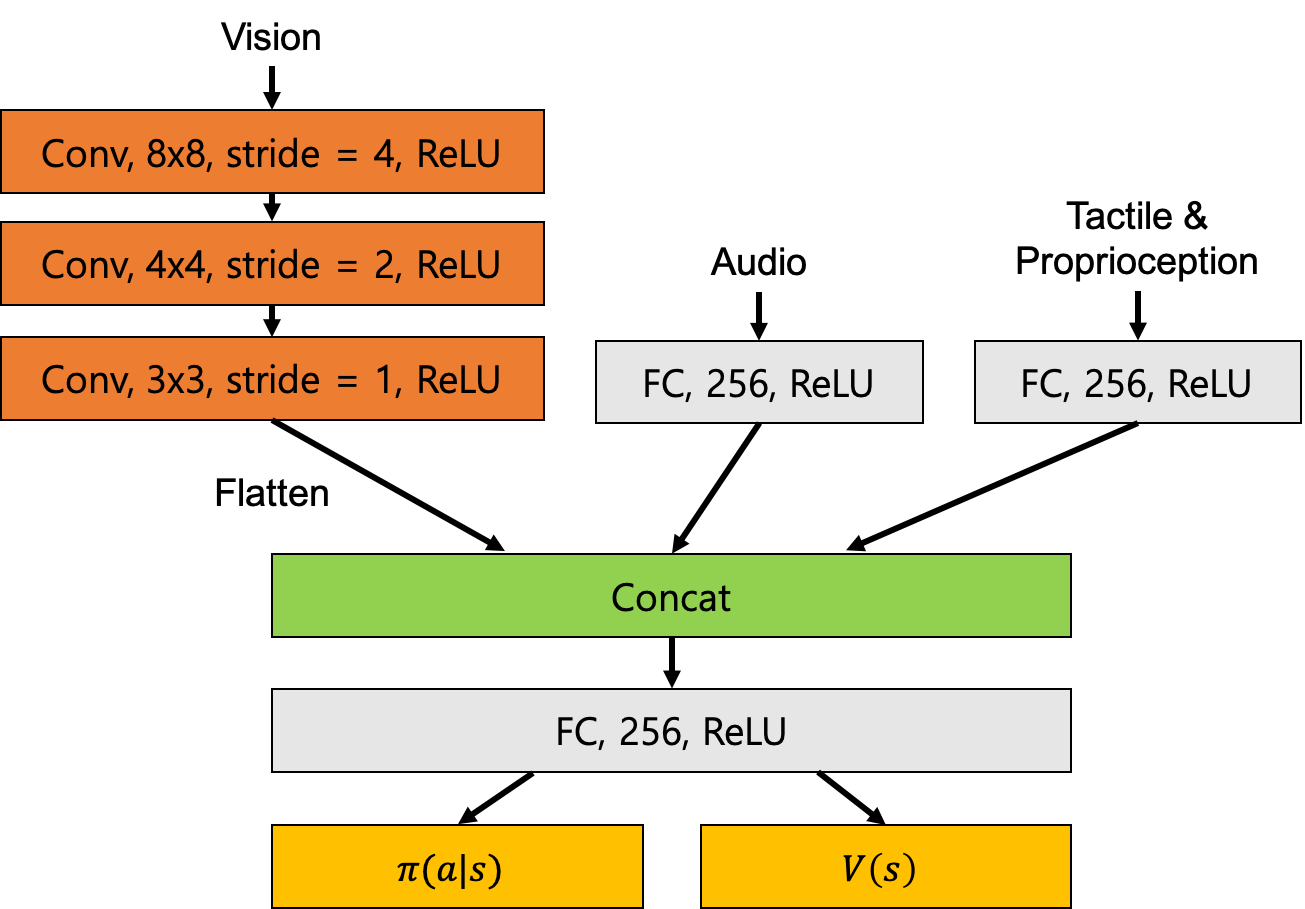}
    \caption{Architecture used for experiments. For SAC, the model outputs the Q value. For supervised learning problems, the model outputs the corresponding results.}
    \label{fig:architecture}
\end{figure}

\subsection{Perception/Action}
\subsubsection{Vision.} 
We sampled binocular RGB vision data in a resolution of 84x84. For the \textit{GrabObject} task, the agent always looks toward the object (i.e., the object is always located at the center of each vision). 

\subsubsection{Audio.} 
We sampled the audio data at the rate of 22050Hz and converted the audio data to frequency-domain by FFT with the window size of 1024. 

\subsubsection{Tactile.}
For the \textit{GrabObject} task, the agent can sense the tactile perception sensed by taxels from bones in hand. We say the taxel is from a bone when the taxel is closer to that bone than any other bones.

\subsubsection{Action.}
For the \textit{GrabObject} task, we set the torque applied to each joint as action and used applyTorque() function to move the agent according to the action. For other tasks, we set the velocity vector of walking as the agent's action and used the walk() function to move the agent. 

\subsection{Helper Rewards}
To accelerate the training process of RL tasks, we gave helper rewards to the agent.

\subsubsection{KickTheBall.} 
We gave a helper reward calculated as $0.01cos\theta$, where $\theta$ is the angle between the velocity vector and the displacement vector to the ball. 

\subsubsection{ObjectNav \& MultiAgentNav.}
We gave a helper reward calculated as $ Vis \cdot (0.05 \cdot a_{f} + 0.03 \cdot a_{l} \cdot L)$. $a_{f}$ and $a_{l}$ is value of velocity vector projected to forward and left direction. $Vis$ is $1$ when the object is visible to the agent. $L$ is $1$ / $-1$ when the object is left / right side of the agent. 

\subsubsection{GrabObject.}
To encourage the agent to move its hands to the object, we gave a helper reward calculated as $d_{t-1} - d_t$, while sum-of-distance $d_t = |R_t - O_t| + |L_t - O_t|$ ($L_t, R_t, O_t$ is the position vector of left hand/right hand/object). On top of that, we also gave a negative quadratic penalty, which is calculated as $-0.004 * |a_t|^2$ where $a_t$ is the action vector, to prevent agent from performing extreme actions.

\subsection{Hyperparameters \& Architectures}

Throughout the experiment, we used the architecture as shown in Fig. \ref{fig:architecture}. For the unused perceptions, the corresponding component is ignored. For the hyper-parameters used in RL algorithms, please see Table \ref{table:param}.

\section{Locomotions with Humanoid Agent}

\begin{figure*}
    \centering
    \includegraphics[width=\textwidth]{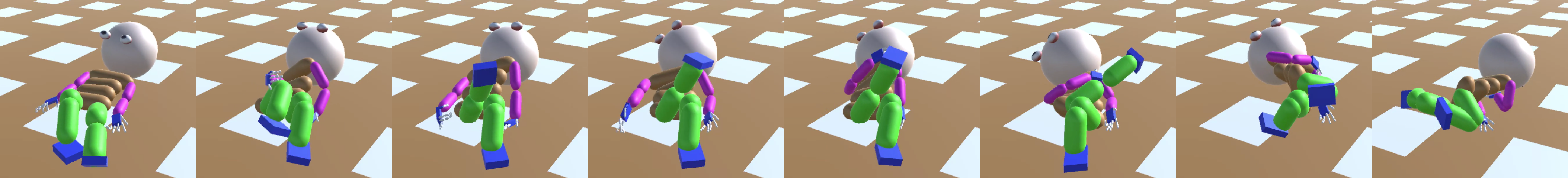}
    \caption{Humanoid agent turning its body.}
    \label{fig:turn}
\end{figure*}

\begin{figure*}
    \centering
    \includegraphics[width=\textwidth]{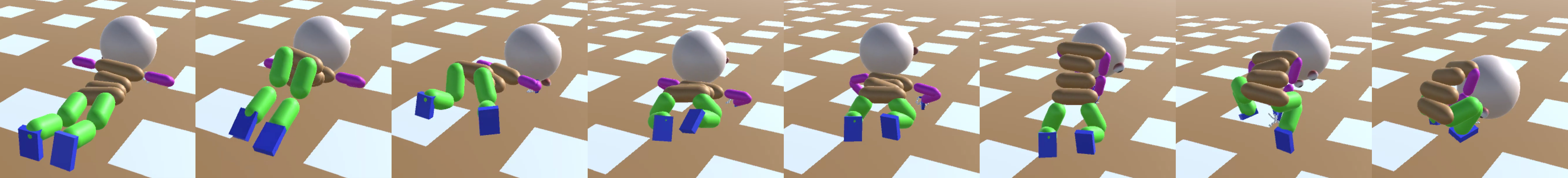}
    \caption{Humanoid agent sitting from lying down.}
    \label{fig:sit}
\end{figure*}

To show the physical capability of provided humanoid agent, we demonstrate locomotions with the humanoid agent: turning its body (Fig. \ref{fig:turn}) and sitting (Fig. \ref{fig:sit}).

\begin{table*}[]
\begin{tabular}{m{0.5\textwidth}m{0.2\textwidth}m{0.2\textwidth}} \hline
\textbf{Parameter}              & \textbf{PPO}                        & \textbf{SAC} \\ \hline
Learning rate                   & Adaptive to KL Div.                 & 2.5e-4       \\ \hline
Number of workers                   & 8                                   & 8            \\ \hline
$\lambda$ (GAE)                 & 0.95                                & N/A          \\ \hline
Clipping ratio                  & 0.2                                 & N/A          \\ \hline
Entropy coefficient             & 0.03                                & 0.01         \\ \hline
Gradient norm clipping          & 5                                   & 5            \\ \hline
$\gamma$ Discount factor        & 0.99                                & 0.99         \\ \hline
Optimizer                       & Adam                                & Adam         \\ \hline
Training epoch/batch per update & 4/4                                 & N/A          \\ \hline
Value function Coefficient      & 0.5                                 & N/A          \\ \hline
Batch size                      & N/A                                 & 64    
\end{tabular}
\caption{Hyperparameters of algorithms used in RL experiments.}
\label{table:param}
\end{table*}

\begin{table*}
\begin{tabular}{|>{\centering\arraybackslash}m{0.15\textwidth}|m{0.35\textwidth}|m{0.4\textwidth}|}
\hline          
\multirow{5}{*}{ObservationUtils}   & float[] getImage (\textit{camera}, \textit{height}, \textit{width}, \textit{grayscale})  & Return the image of the camera with the size of (\textit{height}, \textit{width}). Returns grayscale image if \textit{grayscale} is true, otherwise returns RGB image.         \\ \cline{2-3} 
                                    & float[] getSpatialAudio (\textit{head}, \textit{earL}, \textit{earR}, \textit{env}) & Return the raw spatilized audio data with audio sources included in \textit{env} based on the position of \textit{head} and both ears (\textit{earL}, \textit{earR}). \\ \cline{2-3} 
                                    & float[] getSpatialAudio (\textit{head}, \textit{earL}, \textit{earR}, \textit{env}, Vector3 roomSize, float beta) & Return the raw spatilized audio data, also including the room impulse generated by the room size of \emph{roomSize} and reflection coffecient \emph{beta}. \\ \cline{2-3} 
                                    & float[] getTactile (\textit{mesh})                       & Return the tactile data with the mesh of the agent.\\ \cline{2-3}
                                    & float[] getTactile (\textit{taxels}, \textit{taxelratio}, \textit{debug})   & Return the tactile data with the \textit{taxels} of the agent. If \textit{debug} is true, each taxel displays a blue line which represents the direction and impulse of tactile perception. To prevent massive display of lines during debugging, user could disable some of the taxels using \textit{taxelratio}.
\\ \hline
GeneralAgent                        & void AddObservations (\textit{key}, \textit{observation})      & Add the \textit{observation} vector in its data buffer with its \textit{key}.                                                    
\\ \hline
\end{tabular}
\caption{Example of VECA APIs related to perception of the agent.}
\label{table:APIPerception}
\end{table*}

\begin{table*}
\begin{tabular}{|>{\centering\arraybackslash}m{0.15\textwidth}|m{0.35\textwidth}|m{0.4\textwidth}|}
\hline
Class Name                          & Function                               & Description                                                                  
\\ \hline

\multirow{8}{*}{\begin{tabular}[c]{@{}l@{}}VECAHumanoid\\ExampleInteract\end{tabular}}   
                                    & void walk (\textit{walkSpeed}, \textit{turnSpeed})        & Make the agent walk. Speed and trajectory of the agent is determined by \textit{walkSpeed} and \textit{turnSpeed}.  \\ \cline{2-3} 
                                    & void kick (\textit{obj}) & Make the agent kick the object (\textit{obj}).\\ \cline{2-3} 
                                    & void grab (\textit{obj}) & Make the agent grab the object (\textit{obj}). \\ \cline{2-3} 
                                    & void release () & Make the agent release the grabbed object.  \\ \cline{2-3}                                  
                                    & void adjustFocalLength(\textit{newFocalLength}) & Adjust the focal distance of the cameras.  \\ \cline{2-3}  
                                    & void lookTowardPoint(\textit{pos}) & Make the agent look at 3-dimensional point. Note that this function doesn't rotate the head : it only rotates the camera(eye). Also, the agent will look at the point until releaseTowardPoint() is called, even when the agent is moving.  \\ \cline{2-3}  
                                    & void rotateUpDownHead(\textit{deg}) & Rotate the head in the Up-Down plane. Head goes down when deg$>$0. Please note that this function doesn't have any constraints: Excessive rotation might distort the mesh of the agent.  \\ \cline{2-3} 
                                    & void rotateLeftRightHead(\textit{deg}) & Rotate the head in the Left-right plane. Head goes right when deg$>$0. Please note that this function doesn't have any constraints: Excessive rotation might distort the mesh of the agent.\\ \cline{2-3} 
                                    & void makeSound(\textit{audiodata}) & Make a voice according to the audiodata. Please note that the agent also percieves its voice.\\ \cline{2-3}  
                                    & bool isVisible(\textit{obj}, \textit{cam}) & Check if the object \textit{obj} is visible by the camera \textit{cam}. Note that the object is considered visible even if it is occluded by transparent objects.\\ \cline{2-3} 
                                    & bool isInteractable(\textit{obj}) & Check if the object \textit{obj} is interactable with the agent. The object is considered interactable when 1) the object is visible, 2) there is no transparent objects between the object and the agent 3) the distance is closer than 2.5m.\\ \cline{2-3} 
                                    & List$<$VECAObjectInteract$>$ getInteractableObjects() & Returns the list of objects which is interactable with the agent.\\ \cline{2-3} 
                                    & VECAObjectInteract getInteractableObject() & Returns a single object which is interactable with the agent. If there are multiple interactable objects, it chooses the object which is closest to the center of agent's viewport.
\\ \hline

\multirow{5}{*}{\begin{tabular}[c]{@{}l@{}}VECAHumanoid\\PhysicalExample\\Interact\end{tabular}}  
                                    & List$<$Vector3$>$ GetVelocity() & Returns the velocity of the bones. \\ \cline{2-3} 
                                    & List$<$Vector3$>$ GetAngVelocity() & Returns the angular velocity of the bones.\\ \cline{2-3} 
                                    & List$<$float$>$ GetAngles() & Returns the current angle for all joints (length is same to degree of freedom). \\ \cline{2-3}
                                    & void ApplyTorque(\textit{normalizedTorque}) & Apply the torques to the joint according to the \emph{normalizedTorque}. Input must be the length same to the degree of freedom and within [-1, 1].\\ \cline{2-3}         
                                    & void updateTaxels() & Update the taxels (used in tactile calculation) according to the orientation of the bones and the mesh of the skin. In default, 6 taxels are distributed in triangle formulation for each triangle of the mesh. 
\\ \hline

\end{tabular}
\caption{Example of VECA APIs related to action of the agent.}
\label{table:API}
\end{table*}

\begin{table*}
\begin{tabular}{|>{\centering\arraybackslash}m{0.15\textwidth}|m{0.15\textwidth}|m{0.12\textwidth}|m{0.13\textwidth}|m{0.4\textwidth}|}
\hline  
Figure &
Task          & Observation                     & Action space   &  Description      
\\ \hline
\raisebox{-.5\dimexpr\totalheight-\ht\strutbox}{\includegraphics[width=0.15\textwidth]{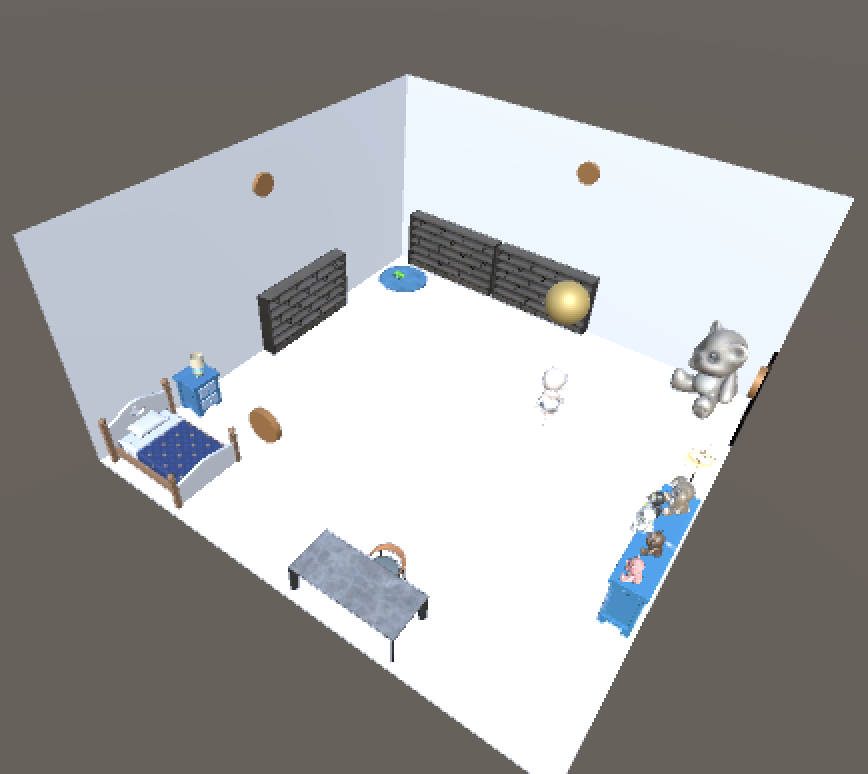}} &
KickTheBall   & Vision, Audio                   & Continuous / Discrete  & 
A ball with a buzzing sound continuously comes out from one of the sidewalls. The agent needs to go closer to the ball to kick it. The agent has to use auditory perception to predict the ball's position since its field of view is limited. 
\\ \hline
\raisebox{-.5\dimexpr\totalheight-\ht\strutbox}{\includegraphics[width=0.15\textwidth]{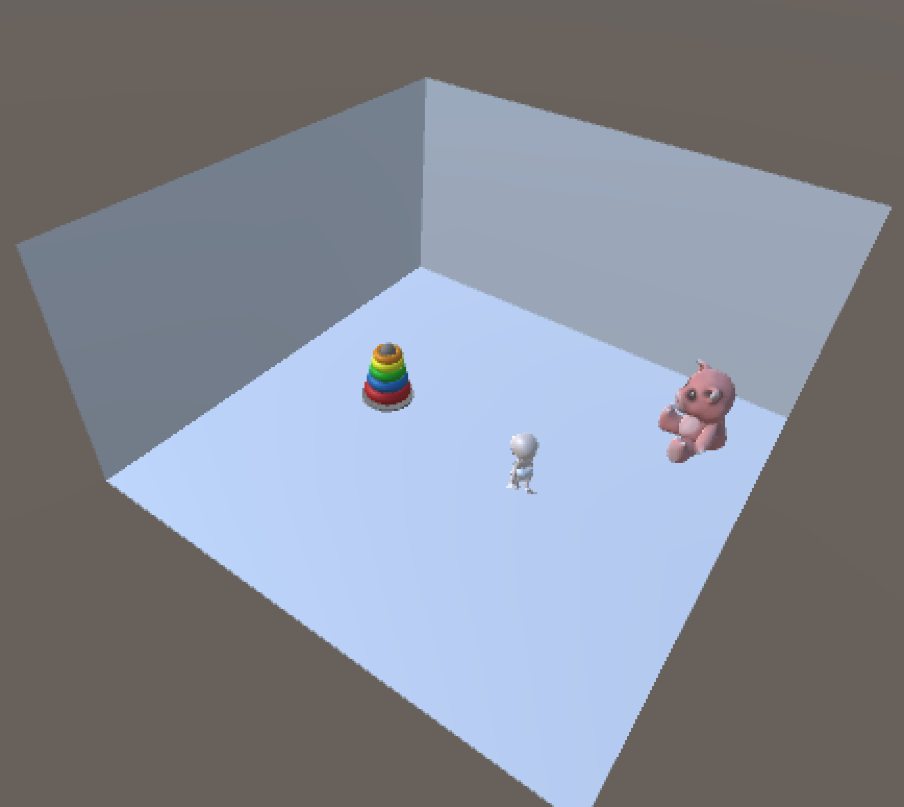}} &
ObjectNav     & Vision                          & Continuous / Discrete  & In the environment, there are multiple objects, including the target object. When the agent has navigated to the target object, then the agent receives +1 reward. If the agent has navigated to the wrong object, then the agent receives -1 reward.
\\ \hline
\raisebox{-.5\dimexpr\totalheight-\ht\strutbox}{\includegraphics[width=0.15\textwidth]{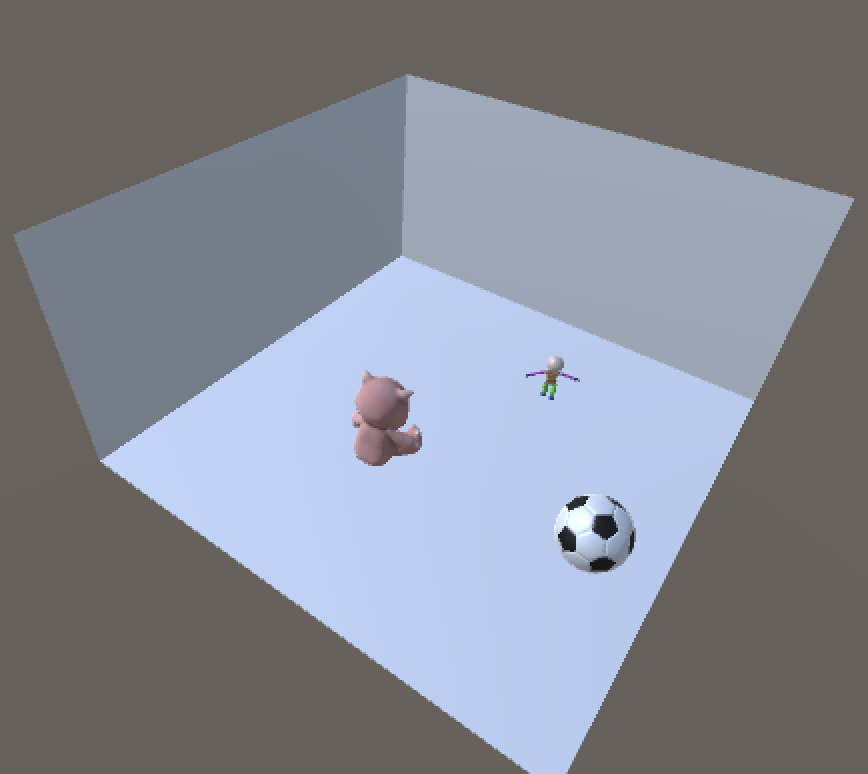}} & 
ObjectPhysNav & Vision, Tactile, Proprioception & Continuous             & Same as \emph{ObjectNav}, but have to learn the task with joint-level actions.
\\ \hline
\raisebox{-.5\dimexpr\totalheight-\ht\strutbox}{\includegraphics[width=0.15\textwidth]{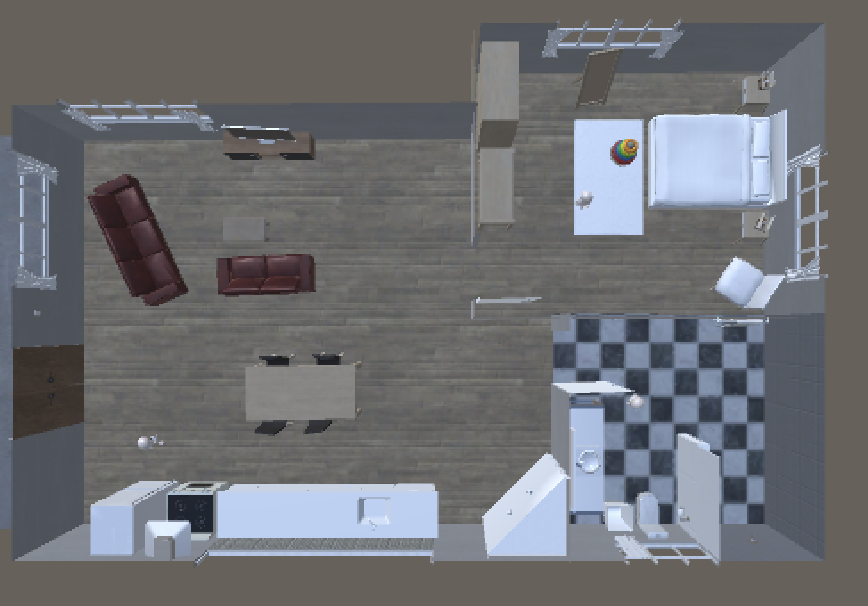}} &
MultiAgentNav & Vision, Tactile, Proprioception & Continuous / Discrete  & Agents are distributed in each room, and the object is in one of those rooms. All agents receive +1 reward when each agent navigates to the object. Without cooperation, the agent must traverse all rooms and find the desired object. The agent who found the object must notify other agents by making a sound, enabling other agents to navigate the object using auditory perception.
\\ \hline
\raisebox{-.5\dimexpr\totalheight-\ht\strutbox}{\includegraphics[width=0.15\textwidth]{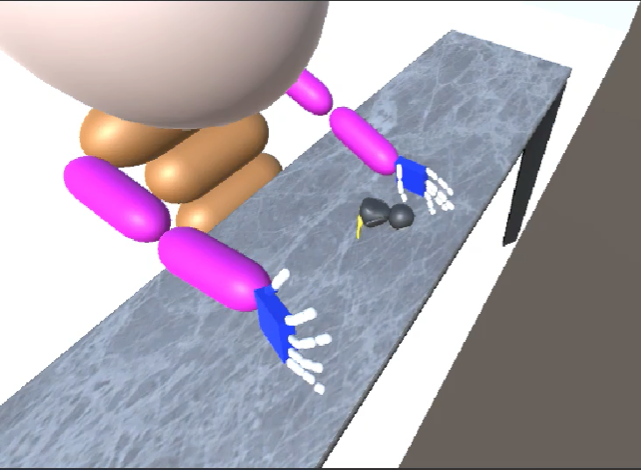}} &
GrabObject    & Vision, Tactile, Proprioception & Continuous             & Learn how to recognize and grab an object with joint-level actions. The agent is rewarded by the vertical position difference when the object is close enough to each hand (preventing the agent from "throwing" the object). 
\\ \hline
\raisebox{-.5\dimexpr\totalheight-\ht\strutbox}{\includegraphics[width=0.15\textwidth]{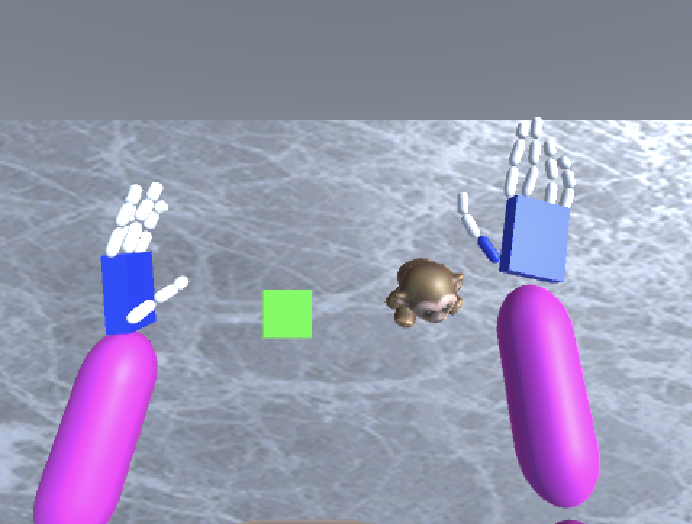}} &
MoveToTarget  & Vision, Tactile, Proprioception & Continuous             & Learn how to recognize and move the object to the target position with joint-level actions. The agent is rewarded by the difference of distance between the target position and the object's position.
\\ \hline
\raisebox{-.5\dimexpr\totalheight-\ht\strutbox}{\includegraphics[width=0.15\textwidth]{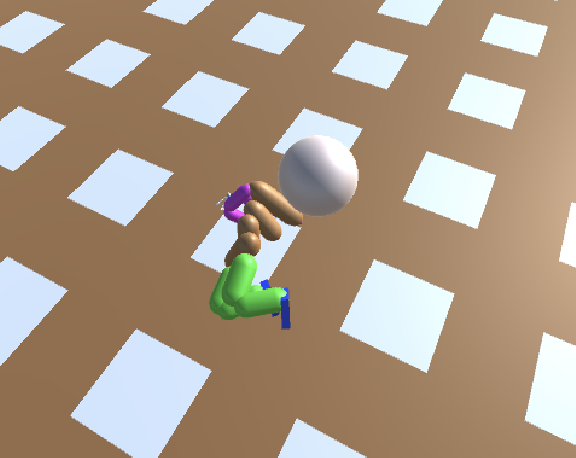}} &
TurnBaby      & Tactile, Proprioception         & Continuous             & Learn how to turn its body with joint-level actions. The agent is rewarded by the orientation of its torso and hip.
\\ \hline
\raisebox{-.5\dimexpr\totalheight-\ht\strutbox}{\includegraphics[width=0.15\textwidth]{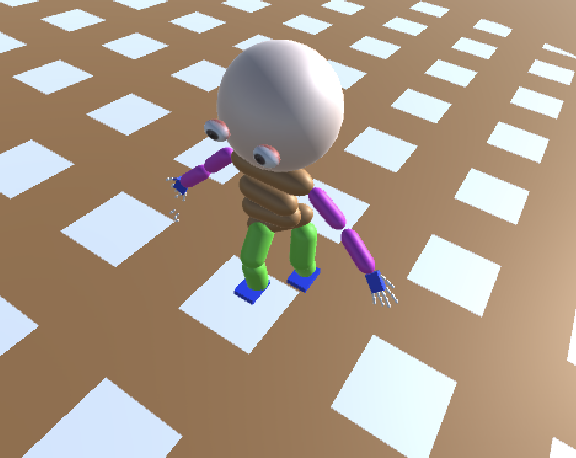}} &
RunBaby       & Tactile, Proprioception         & Continuous             & Learn how to run with joint-level actions. The agent is rewarded by the velocity to the direction of its torso.
\\ \hline
\end{tabular}
\end{table*}

\begin{table*}
\begin{tabular}{|>{\centering\arraybackslash}m{0.15\textwidth}|m{0.15\textwidth}|m{0.12\textwidth}|m{0.13\textwidth}|m{0.4\textwidth}|}
\hline  
\raisebox{-.5\dimexpr\totalheight-\ht\strutbox}{\includegraphics[width=0.15\textwidth]{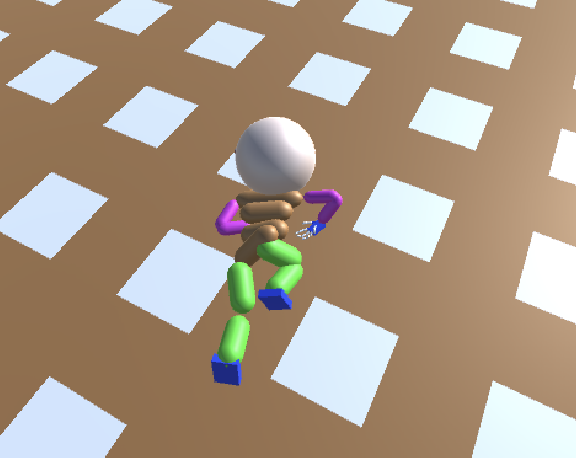}} &
CrawlBaby     & Tactile, Proprioception         & Continuous             & Learn how to crawl with joint-level actions. The agent is rewarded by the velocity and the orientation of its torso and hip.
\\ \hline
\raisebox{-.5\dimexpr\totalheight-\ht\strutbox}{\includegraphics[width=0.15\textwidth]{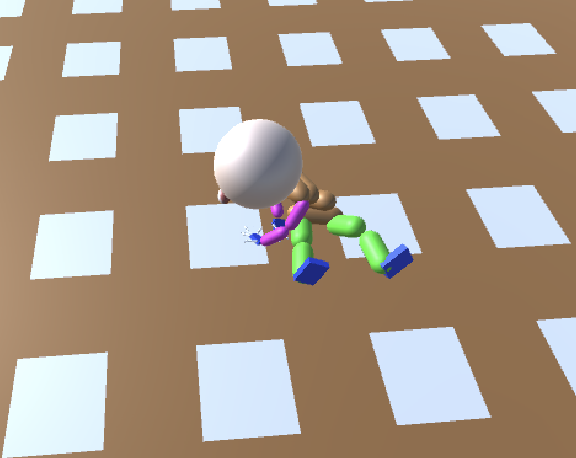}} &
SitBaby       & Tactile, Proprioception         & Continuous             & Learn how to sit with joint-level actions. The agent is rewarded by the position of its torso and head.
\\ \hline
\raisebox{-.5\dimexpr\totalheight-\ht\strutbox}{\includegraphics[width=0.15\textwidth]{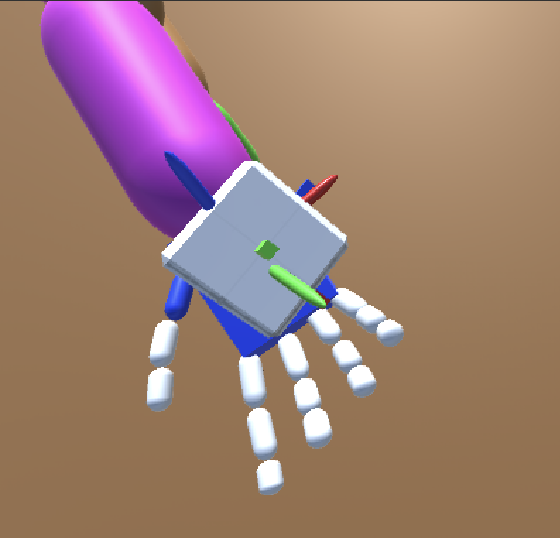}} &
RotateCube    & Tactile, Proprioception         & Continuous             & Learn how to rotate the cube to target rotation using its hand with joint-level actions. The agent is rewarded by the difference in distance between the target orientation and the cube's orientation.
\\ \hline
\raisebox{-.5\dimexpr\totalheight-\ht\strutbox}{\includegraphics[width=0.15\textwidth]{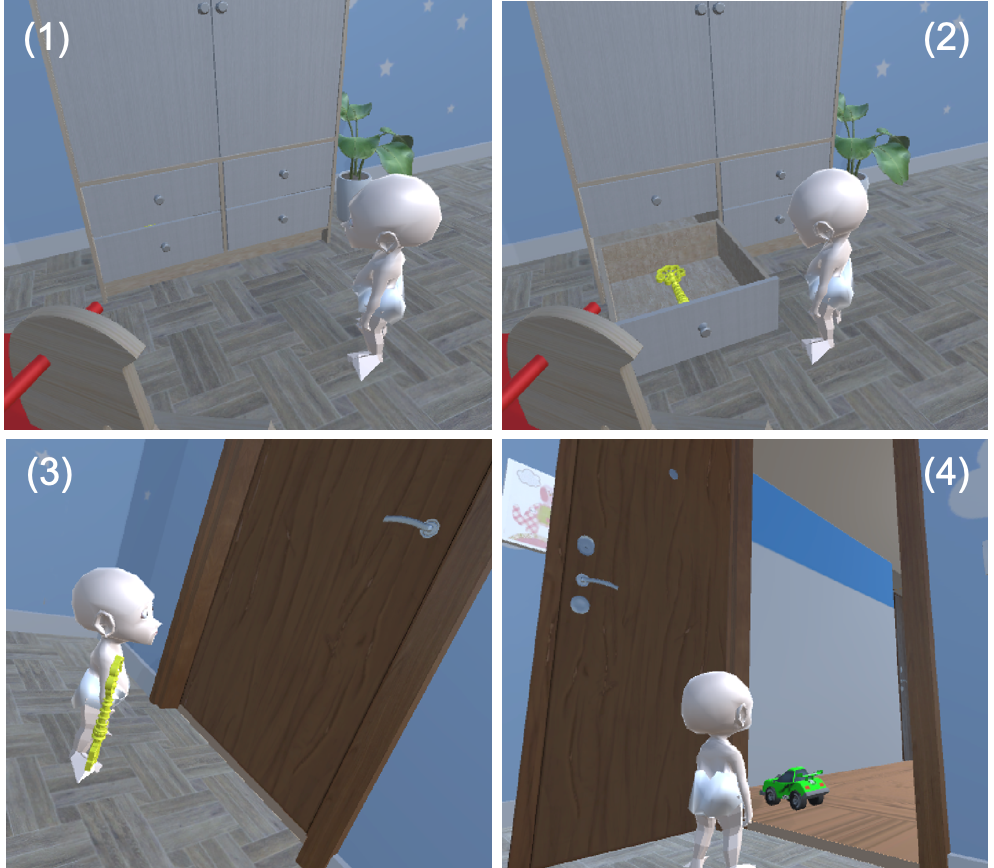}} &
Babyzuma's revenge  & Vision, Audio             & Continuous \& Discrete & Similar to \emph{Montezuma's revenge}, the agent has to learn complex sequence of interactions to complete the level in sparse reward settings, with vision and audio senses. For instance, the agent has to: 1) navigate to the drawer, 2) open the drawer, 3) grab the key (in the drawer), 4) open the door to clear the first room.
\\ \hline
\raisebox{-.5\dimexpr\totalheight-\ht\strutbox}{\includegraphics[width=0.15\textwidth]{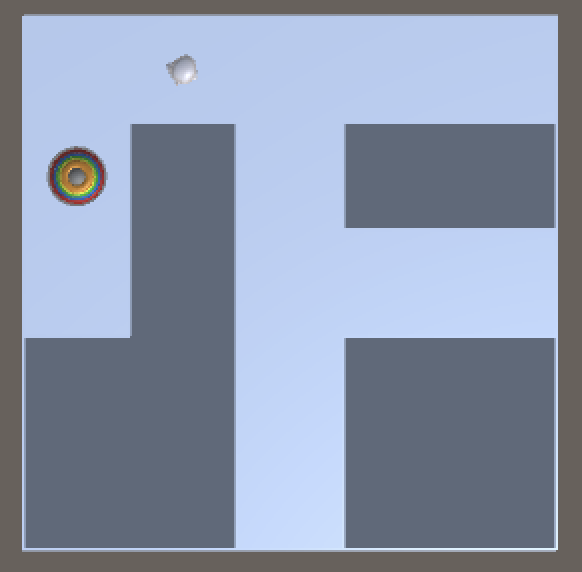}} &
MazeNav       & Vision                          & Continuous / Discrete  & Adopted the navigation experiment from RL$^2$\cite{RL2}. The agent is rewarded when the agent navigates to the object. The agent can repeat five trials for each maze.
\\ \hline
\raisebox{-.5\dimexpr\totalheight-\ht\strutbox}{\includegraphics[width=0.15\textwidth]{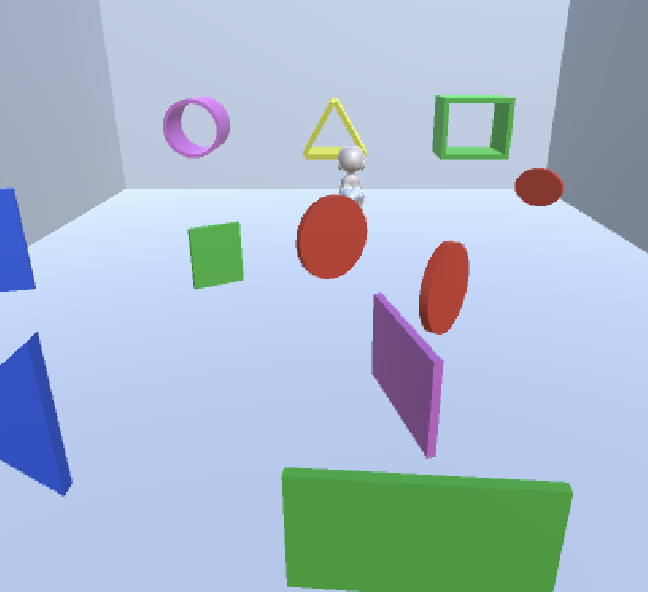}} &
ShapeSort     & Vision                          & Continuous \& Discrete  & The agent has to sort the objects according to their shape. The agent is rewarded when the agent grabs the objects and puts the object to the correct shape basket.
\\ \hline
\raisebox{-.5\dimexpr\totalheight-\ht\strutbox}{\includegraphics[width=0.15\textwidth]{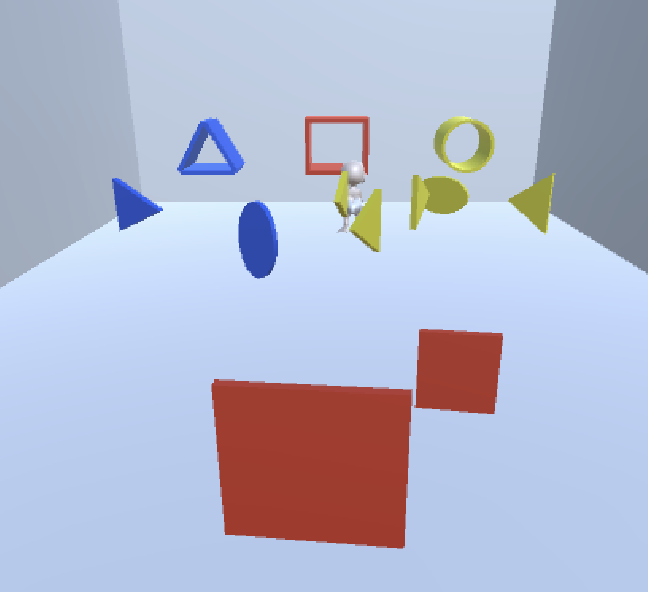}} &
ColorSort     & Vision                          & Continuous \& Discrete  & The agent has to sort the objects according to their color. The agent is rewarded when the agent grabs the objects and puts the object to the correct color basket.
\\ \hline
\raisebox{-.5\dimexpr\totalheight-\ht\strutbox}{\includegraphics[width=0.15\textwidth]{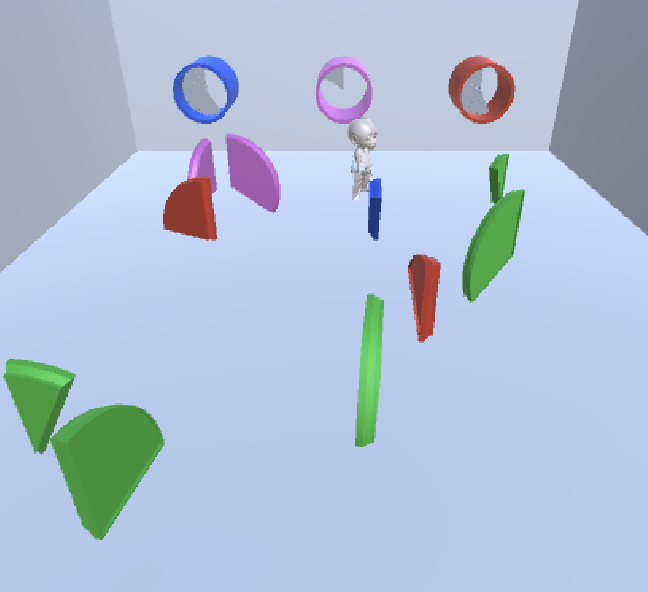}} &
FillFraction  & Vision                          & Continuous \& Discrete  &  The agent has to fill three circles using the fraction circle objects. The agent can not put the object when the sum of the fraction exceeds the circle. The agent is rewarded when the agent puts the object to the circle and is additionally rewarded when it is full.
\\ \hline
\end{tabular}
\end{table*}

\begin{table*}
\begin{tabular}{|>{\centering\arraybackslash}m{0.15\textwidth}|m{0.15\textwidth}|m{0.12\textwidth}|m{0.13\textwidth}|m{0.4\textwidth}|}
\hline  
\raisebox{-.5\dimexpr\totalheight-\ht\strutbox}{\includegraphics[width=0.15\textwidth]{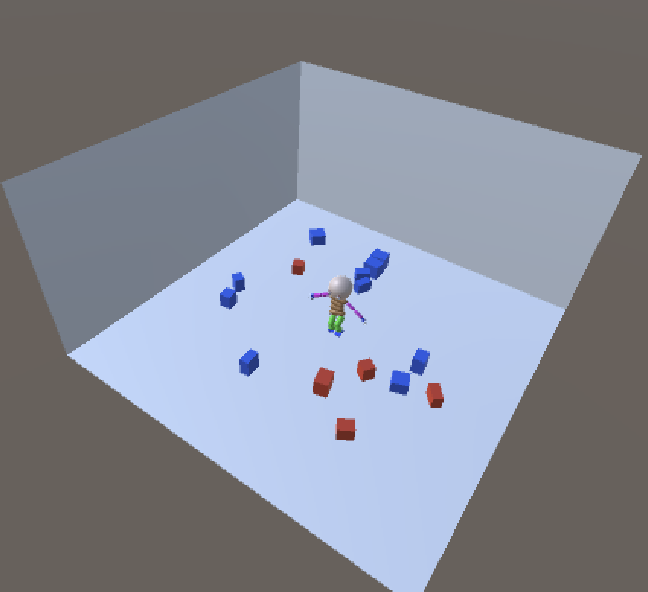}} &
PileUpBlock   & Vision, Tactile, Proprioception & Continuous              & The agent has to pile up the (red) blocks with joint-level actions. The agent is rewarded according to the sum of the y-axis position of red blocks.
\\ \hline
\raisebox{-.5\dimexpr\totalheight-\ht\strutbox}{\includegraphics[width=0.15\textwidth]{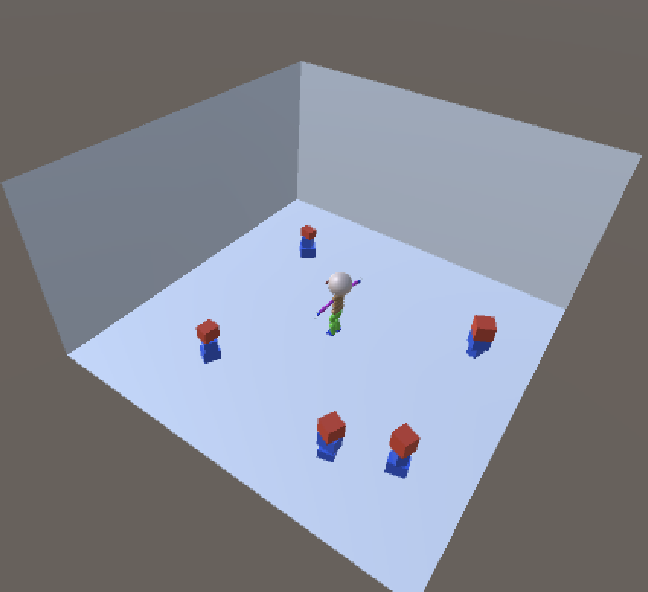}} &
PutDownBlock  & Vision, Tactile, Proprioception & Continuous              & The agent has to put down (only the red) blocks with joint-level actions. The agent is rewarded according to the negative sum of the y-axis position of red blocks. Additionally, the agent gets a negative reward when the blue blocks fall. 
\\ \hline
\raisebox{-.5\dimexpr\totalheight-\ht\strutbox}{\includegraphics[width=0.15\textwidth]{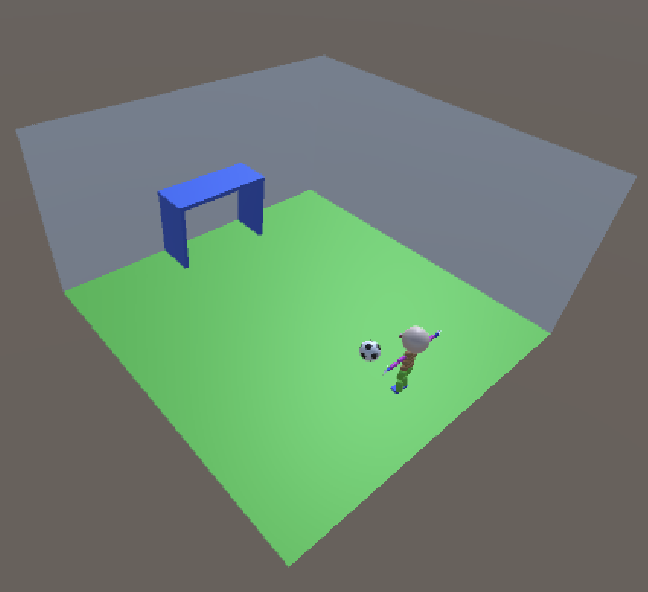}} &
SoccerOnePlayer & Vision, Tactile, Proprioception & Continuous             & Put the ball into the goal with joint-level actions. Note that there is no rule: e.g., the agent may use hands.
\\ \hline
\raisebox{-.5\dimexpr\totalheight-\ht\strutbox}{\includegraphics[width=0.15\textwidth]{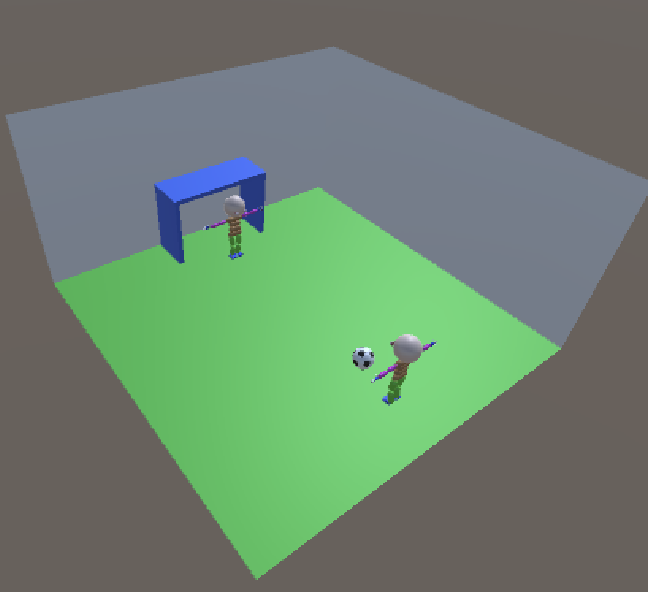}} &
SoccerTwoPlayer & Vision, Tactile, Proprioception & Continuous             & The attacker must put the ball into the goal, and the defender must block the ball from the goal with joint-level actions. Note that there is no rule: e.g., the agent may use hands and block or tackle the opponent player.
\\ \hline
\raisebox{-.5\dimexpr\totalheight-\ht\strutbox}{\includegraphics[width=0.15\textwidth]{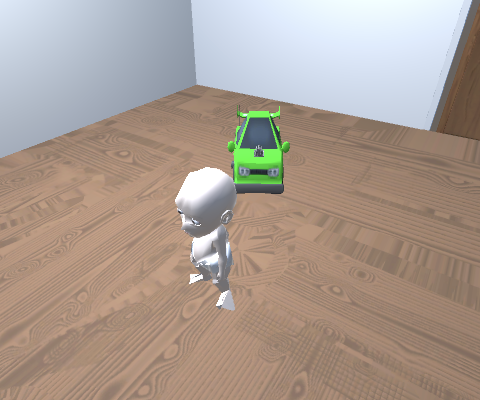}} &
DodgeCar        & Vision                          & Continuous / Discrete  & The toy car moves around in the room fast. The agent must dodge the car.
\\ \hline
\raisebox{-.5\dimexpr\totalheight-\ht\strutbox}{\includegraphics[width=0.15\textwidth]{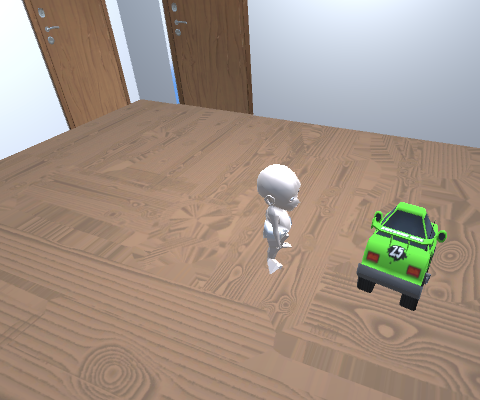}} &
ChaseCar        & Vision, Tactile, Proprioception & Continuous \& Discrete & The toy car moves around in the room fast. The agent must chase and interact with the car, but must not collide with the car.
\\ \hline
\raisebox{-.5\dimexpr\totalheight-\ht\strutbox}{\includegraphics[width=0.15\textwidth]{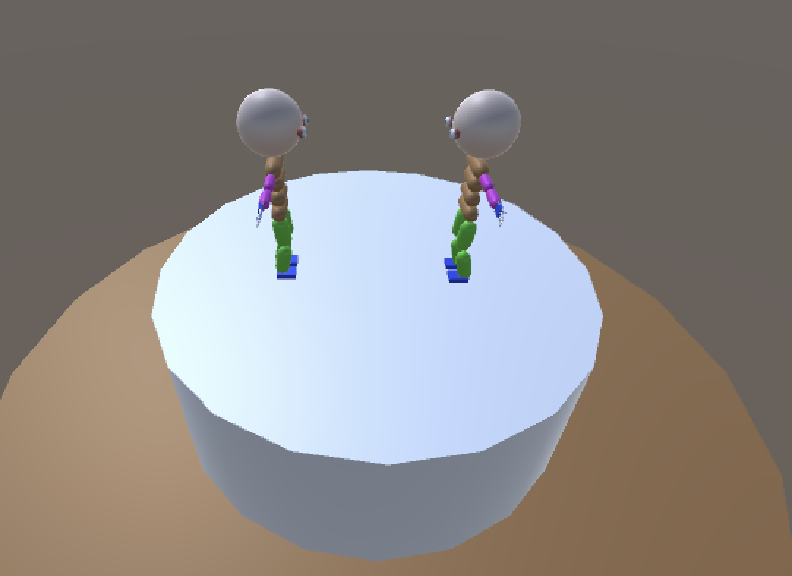}} &
PushOut1v1      & Vision, Tactile, Proprioception & Continuous             & The agent must push other agents out of the ring to win. The agent is equipped with joint-level actions.
\\ \hline
\raisebox{-.5\dimexpr\totalheight-\ht\strutbox}{\includegraphics[width=0.15\textwidth]{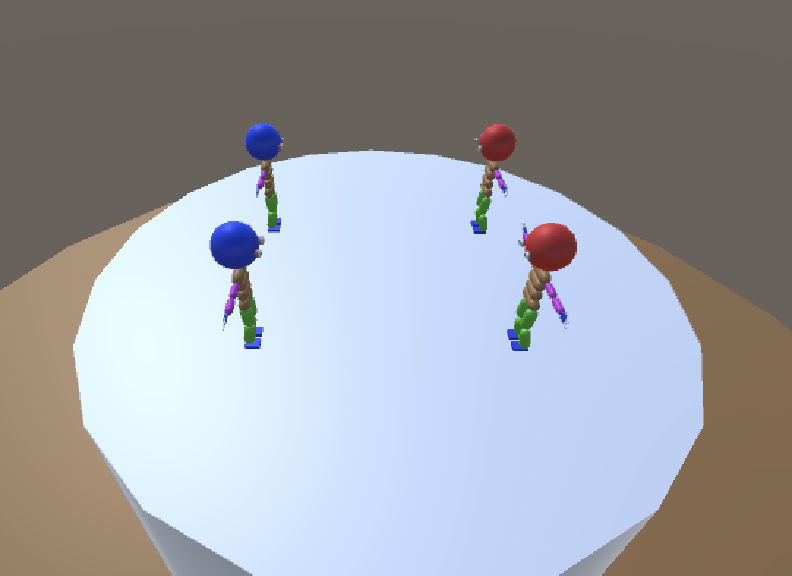}} &
PushOut2v2      & Vision, Tactile, Proprioception & Continuous             & Agents must push agents of the opponent team out of the ring to win. The agent is equipped with joint-level actions.
\\ \hline
\end{tabular}
\caption{Description of various tasks provided by VECA.}
\end{table*}

\end{document}